\documentclass[journal,10pt,twocolumn]{IEEEtran}

\usepackage{amsmath,amssymb,amsfonts}
\usepackage{graphicx}
\usepackage{booktabs}
\usepackage{algorithm}
\usepackage{algpseudocode}
\usepackage{cite}
\usepackage{url}
\usepackage{microtype}
\usepackage{multirow}
\usepackage{placeins}
\usepackage{dblfloatfix}
\usepackage[hidelinks]{hyperref}

\begin{document}

\title{ClickGuard: A Trustworthy Adaptive Fusion Framework for Clickbait Detection}

\author{Naman~Chawla,
        Chhavi~Dhiman,
        Riya~Dhami,
        Gaurav~Kumar,
        and~Ganesh~Naik%
\thanks{N. Chawla and C. Dhiman contributed equally to this work (co-first authors).}%
\thanks{C. Dhiman and G. Kumar are with the Department of Electronics and Communication Engineering, Delhi Technological University, Delhi, India.}%
\thanks{N. Chawla is with the Department of Computer Science \& Engineering, Bhagwan Parshuram Institute of Technology, Delhi, India.}%
\thanks{R. Dhami is with the Department of Electronics \& Communication Engineering, Bhagwan Parshuram Institute of Technology, Delhi, India.}%
\thanks{G. Naik is with the College of Medicine \& Public Health, Flinders University, Bedford Park, SA-5042, Australia.}%
\thanks{Corresponding authors: \url{namanchawla2404@gmail.com}, \url{chhavi.dhiman@dtu.ac.in}}%
\thanks{Source code and implementation artifacts are publicly available at \url{https://github.com/palindromeRice/ClickGuard}.}%
}

\maketitle

\begin{abstract}
The widespread proliferation of clickbait headlines, crafted to mislead readers and artificially maximize online user engagement, poses a severe challenge to digital content integrity. These headlines exploit sensationalism, forward referencing, ambiguous phrasing, and emotional triggers, underscoring the urgent need for robust automated detection. In this paper, we propose \textbf{ClickGuard}: a trustworthy adaptive fusion framework for clickbait detection. ClickGuard integrates contextual transformer embeddings from BERT with hand-crafted syntactic and structural linguistic features via a novel Syntactic-Semantic Adaptive Fusion Block (SSAFB). The architectural pipeline introduces a hybrid 1D-CNN and BiLSTM network to model local and sequential patterns from tabular linguistic features, coupled with a Multi-Head Attention (MHA) module to enrich contextual semantic representations. Comprehensive evaluations across three public benchmark datasets demonstrate the effectiveness of our framework, achieving 96.93\% validation accuracy on Dataset 1, 94.39\% on Dataset 2, and 92.05\% on Dataset 3. We further conduct rigorous model interpretability analysis using Permutation Feature Importance (PFI) and Local Interpretable Model-agnostic Explanations (LIME), alongside perturbation robustness checks across typos, synonym replacement, word shuffling, and deletion. Extensive ablation experiments confirm the critical role of adaptive fusion weighting in optimizing cross-modal feature interactions.
\end{abstract}


\section{Introduction}
\IEEEPARstart{T}{he} exponential growth of digital media and the intense commercial race for user engagement have fueled the prevalence of ``clickbait'' headlines~\cite{ref1}. Clickbait refers to sensationalized or misleading snippets designed to pique curiosity, artificially driving traffic and ad revenue~\cite{ref2}. By employing exaggerated claims, forward references, or emotional triggers, clickbait misrepresents underlying articles, degrading user trust and exacerbating online misinformation~\cite{ref3}.

Early clickbait mitigation relied on traditional machine learning algorithms with handcrafted linguistic indicators, such as punctuation frequency, POS distributions, and sentiment polarity~\cite{ref4, ref6, ref12}. While computationally efficient, these models struggled to capture deeper semantic dependencies and subtle syntactic structures~\cite{ref18}. Subsequent deep learning models adopted Convolutional Neural Networks (CNNs)~\cite{ref13, ref14} and Recurrent Neural Networks (RNNs/LSTMs)~\cite{ref9, ref18} to automate representation learning. However, sequential models often lose long-range context, whereas pure language models like BERT may overlook domain-specific stylistic conventions evident in headline structure~\cite{ref20, ref21}. While multimodal methods have integrated thumbnails and video metadata~\cite{ref7, ref16, ref17}, text-based headlines remain the dominant format across social feeds, demanding superior linguistic modeling.

To resolve these challenges, we introduce \textbf{ClickGuard}, a dual-pathway architecture featuring a Syntactic-Semantic Adaptive Fusion Block (SSAFB). The major contributions of this work are summarized as follows:
\begin{enumerate}
    \item We propose \textbf{ClickGuard}, an end-to-end framework combining pre-trained BERT contextual embeddings with an 18-dimensional structural feature set via a hybrid CNN-BiLSTM structural encoder and a multi-head attention contextual module.
    \item We formulate the \textbf{Syntactic-Semantic Adaptive Fusion Block (SSAFB)}, which dynamically calculates non-linear gated weights ($\alpha, 1-\alpha$) to blend contextual and structural representations adaptively.
    \item We conduct thorough trustworthiness and interpretability studies using Permutation Feature Importance (PFI) and LIME, visualizing global feature hierarchies and local instance-level attributions.
    \item We execute comprehensive perturbation robustness stress-tests across typos, omissions, and lexical substitutions, validating model stability under input noise across three standard benchmark datasets.
\end{enumerate}

The remainder of this paper is structured as follows: Section~II reviews related work. Section~III details the ClickGuard architecture and SSAFB design. Section~IV presents empirical results, ablation studies, interpretability, and robustness evaluations. Section~V concludes with future research directions.

\section{Related Work}

\subsection{Syntactic and Semantic Feature Learning}
Linguistic features play a vital role in natural language processing. Syntactic attributes encapsulate grammatical structures, while semantic features capture contextual meaning. Liu et al.~\cite{ref20} proposed a WeChat clickbait detector combining BERT, BiLSTM, and Graph Attention Networks. Rony et al.~\cite{ref18} developed an LSTM framework using POS analysis, reaching 97\% accuracy. Pujahari and Sisodia~\cite{ref19} employed word vector similarity with t-SNE clustering, showing that single categorization techniques are insufficient. Thakur et al.~\cite{ref22} utilized RNNs for syntactic dependency modeling. Elayashar et al.~\cite{ref3} demonstrated that handcrafted syntactic features remain indispensable for classifying sensationalism. Meng et al.~\cite{ref23} developed the MCBD model using gated convolutions and deep relevance matching, while Ma et al.~\cite{ref24} achieved 98.42\% accuracy leveraging 18 lexicon- and format-based features.

\subsection{Attention-Based Feature Learning}
Attention mechanisms allow models to dynamically weight salient tokens. Wei and Nguyen~\cite{ref10} proposed KED using WordNet for semantic guidance, outperforming standard pretrained models. Sharma and Patel~\cite{ref1} fused attention with regression techniques to boost web content classification. Yi et al.~\cite{ref25} proposed Contrastive Variational Modeling (CVM) for simultaneous clickbait headline generation and detection.

\subsection{Graph Neural Networks and Structural Representations}
Graph Neural Networks (GNNs) effectively model non-linear relational structures. Zhang et al.~\cite{ref26} developed the Recurrent Graph Feature Network (RGFN) on heterogeneous text graphs. Do et al.~\cite{ref27} combined shallow and deep semantic representations with GNNs for fake news detection. Wang et al.~\cite{ref28} applied graph convolutions within a counterfactual framework to mitigate clickbait exposure bias.

\subsection{Generative and Transfer Models}
Generative Adversarial Networks (GANs) have advanced style transfer between factual and sensational text. Agarwal and Kundu~\cite{ref29} proposed StyleTransformer with CycleGAN architectures for unsupervised clickbait style transfer while preserving underlying factual meaning.

\section{Proposed Methodology}
The clickbait identification task is formulated as a binary classification problem. Let $\mathcal{H}$ denote the headline corpus, and let $\mathcal{T} = \{(h_i, y_i)\}_{i=1}^N$ represent the labeled training set, where $h_i \in \mathcal{H}$ is the headline text and $y_i \in \{0, 1\}$ indicates the ground truth ($0 = \text{non-clickbait}, 1 = \text{clickbait}$). We seek to optimize the parameters $\theta^*$ of a mapping function $f(h_i; \theta) \to \hat{y}_i \in [0, 1]$ minimizing the binary cross-entropy loss:
\begin{equation}
\theta^* = \arg\min_{\theta} \sum_{i=1}^N \mathcal{L}_{\text{BCE}}(f(h_i; \theta), y_i)
\label{eq:opt}
\end{equation}
where:
\begin{equation}
\mathcal{L}_{\text{BCE}}(\hat{y}_i, y_i) = -\left[ y_i \log \hat{y}_i + (1 - y_i) \log(1 - \hat{y}_i) \right]
\end{equation}

\begin{figure*}[t]
    \centering
    \includegraphics[width=0.82\textwidth]{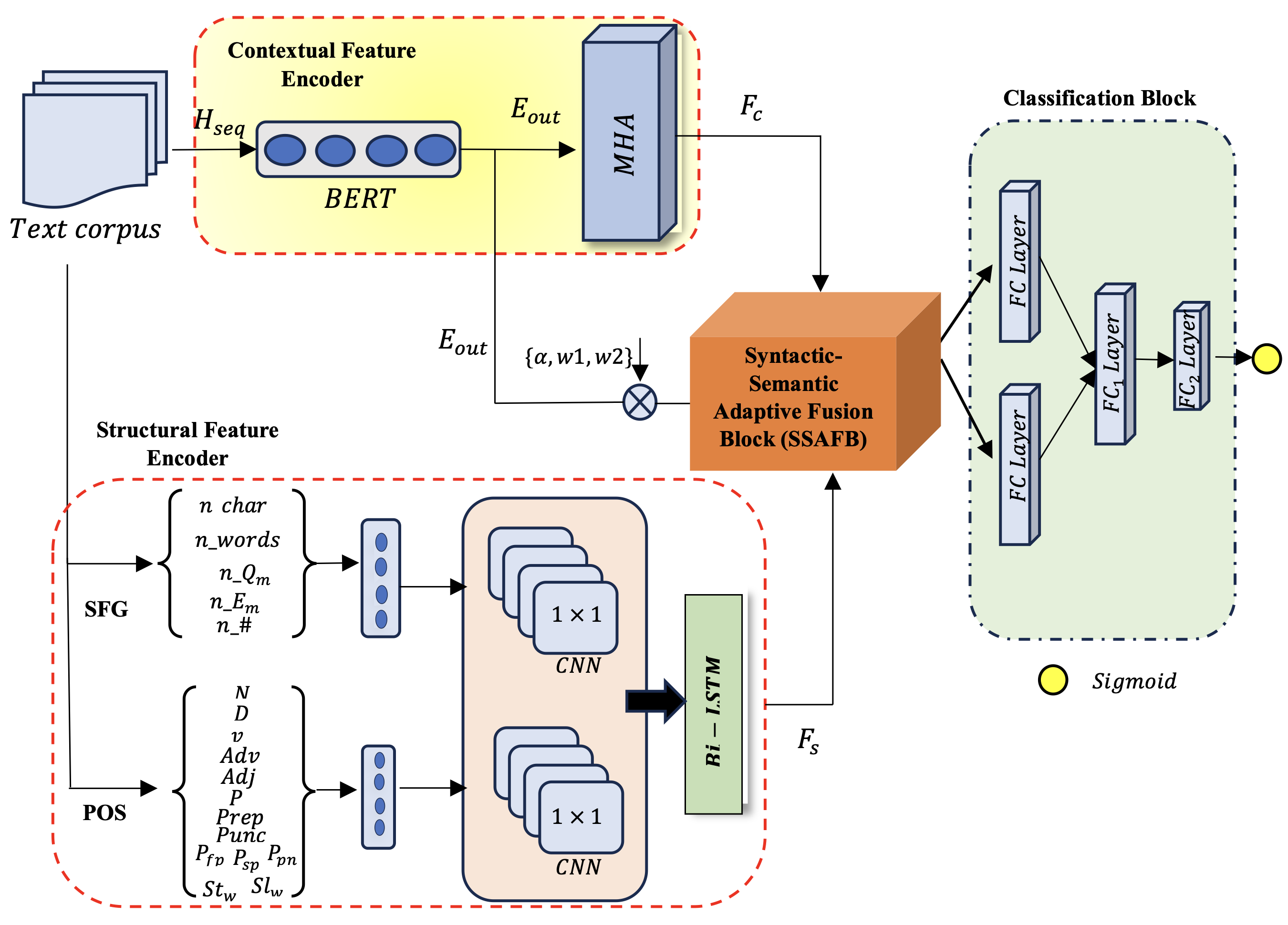}
    \caption{Proposed ClickGuard Architecture: Contextual Feature Encoder (BERT + MHA), Structural Feature Encoder (1D-CNN + BiLSTM), and Syntactic-Semantic Adaptive Fusion Block (SSAFB).}
    \label{fig:architecture}
\end{figure*}

As illustrated in Fig.~\ref{fig:architecture}, ClickGuard operates across three stages: (1) dual-encoder representation learning, (2) adaptive feature fusion via SSAFB, and (3) classification.

\subsection{Linguistic Structural Feature Extraction}
\label{sec:features}
To capture stylistic hallmarks of clickbait, we extract an 18-dimensional structural feature vector $S \in \mathbb{R}^{18}$ combining Sentence Feature Generation (SFG) and Part-of-Speech (POS) metrics, detailed in Table~\ref{tab:features}. 

\begin{table}[t]
\centering
\caption{Description of Structural Feature Set ($S$)}
\label{tab:features}
\begin{tabular}{@{}lll@{}}
\toprule
\textbf{Category} & \textbf{Notation} & \textbf{Extracted Features} \\ \midrule
POS (13) & $P_{fp}, P_{sp}, P_{pn}$ & 1st, 2nd, possessive pronouns \\
         & $N, V, Adj, Adv$         & Nouns, verbs, adjectives, adverbs \\
         & $P, Prep, Punc$          & Pronouns, prepositions, punctuations \\
         & $D, St_w, Sl_w$          & Determiners, stopwords, slang words \\ \midrule
SFG (5)  & $n_{\text{char}}, n_{\text{words}}$ & Character and word counts \\
         & $n_{Qm}, n_{Em}, n_{\#}$           & Question, exclamation marks, hashtags \\ \bottomrule
\end{tabular}
\end{table}

\begin{figure}[t]
    \centering
    \includegraphics[width=\columnwidth]{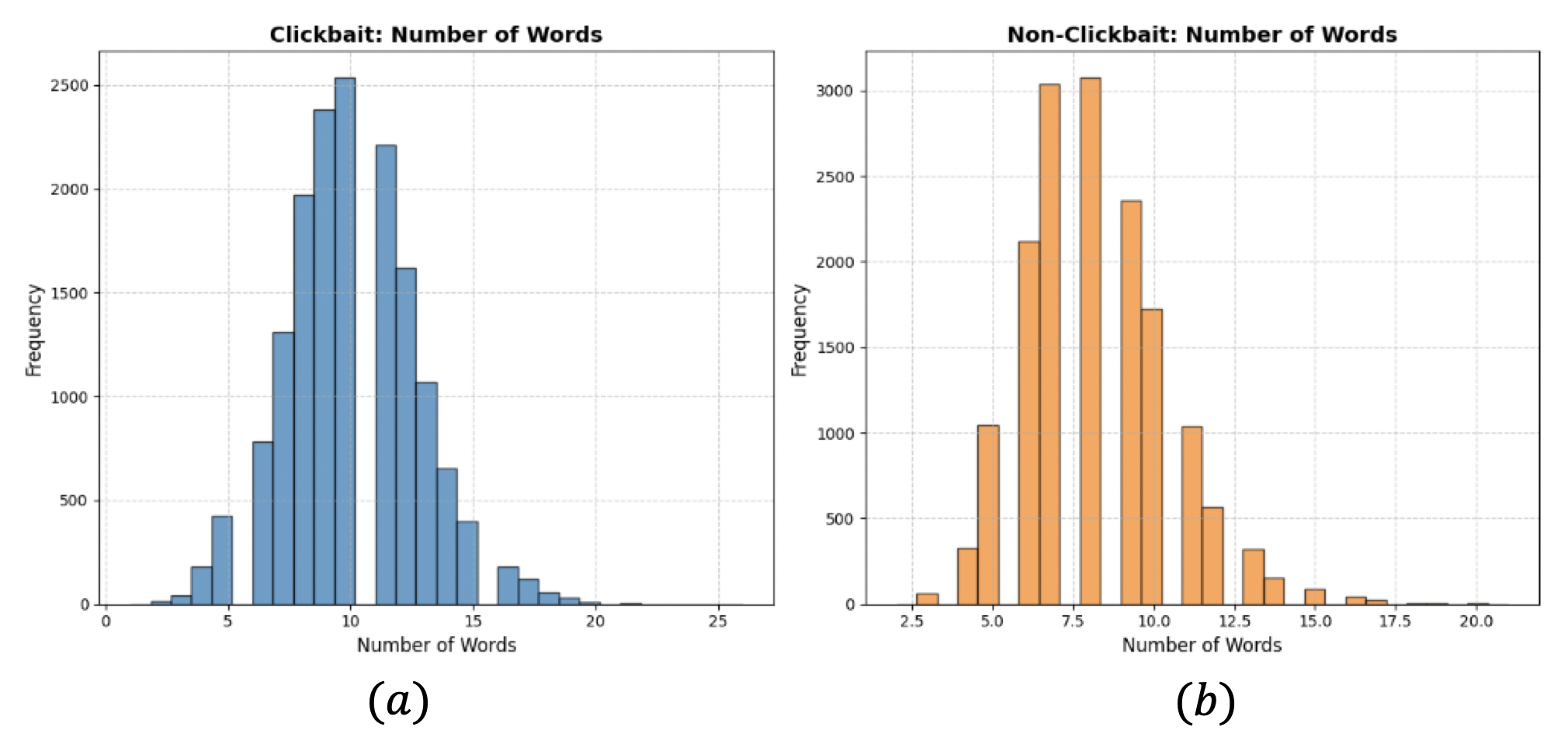}
    \caption{Histograms of structural features for (a) Clickbait and (b) Non-Clickbait headlines from Dataset 1.}
    \label{fig:histograms}
\end{figure}

\begin{figure}[t]
    \centering
    \includegraphics[width=0.80\columnwidth]{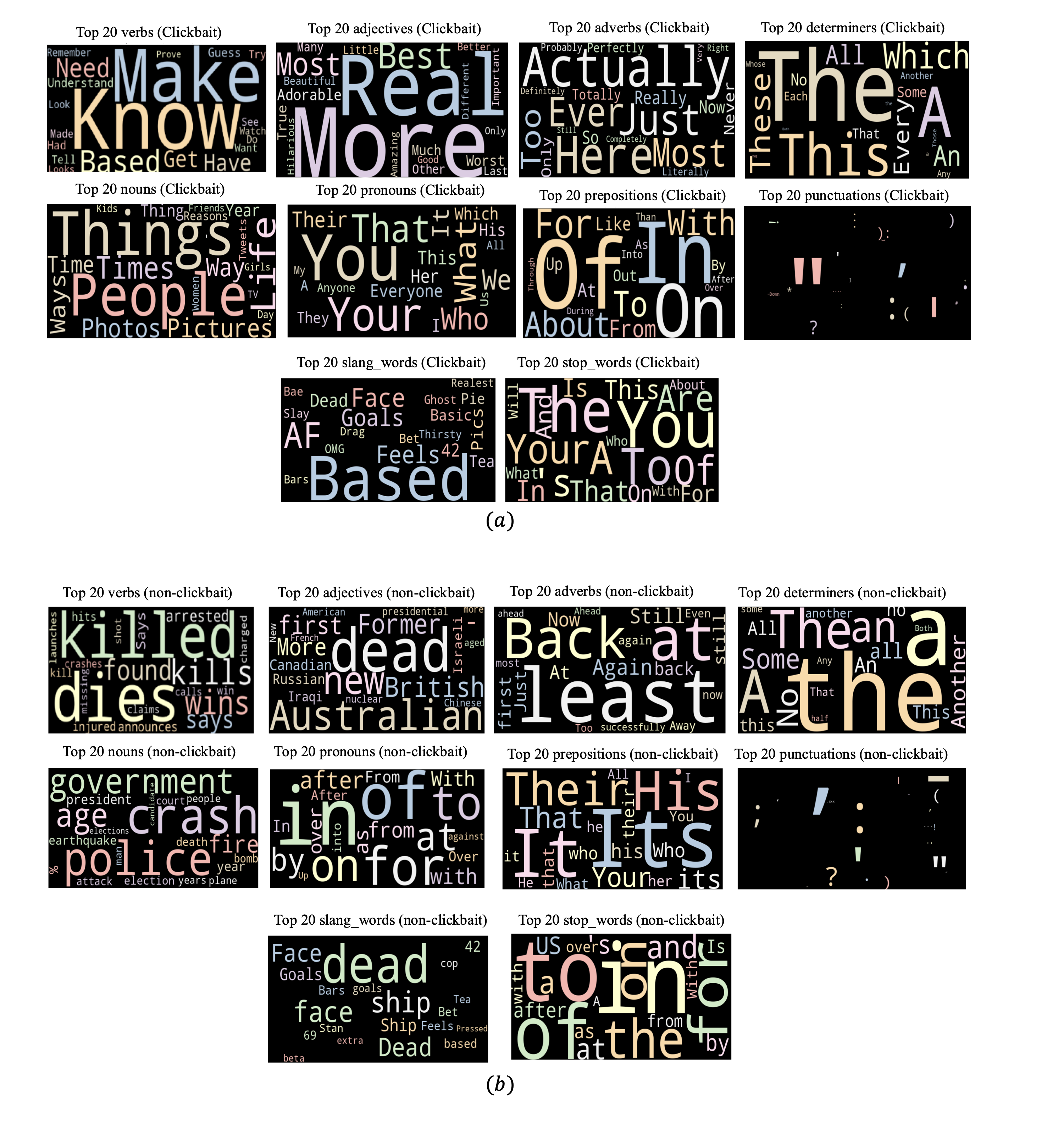}
    \caption{Top-20 word clouds for (a) Clickbait and (b) Non-Clickbait linguistic features from Dataset 1.}
    \label{fig:wordclouds}
\end{figure}

Fig.~\ref{fig:histograms} depicts the distinct length distributions: clickbait peaks around 55--60 characters and 8 words in a normal distribution, whereas non-clickbait is right-skewed with shorter headlines (40 characters, 5--7 words). Fig.~\ref{fig:wordclouds} shows word clouds illustrating that clickbait heavily employs direct second-person pronouns (``you'', ``your''), conversational verbs (``know'', ``make''), and slang terms.

\subsection{Contextual Feature Encoder}
The input text $h_i$ is tokenized into input IDs, attention masks, and token type IDs:
\begin{equation}
H_{\text{seq}} = \text{BERTTokenizer}(h_i) \in \mathbb{R}^{L}
\end{equation}
The tokens pass through pre-trained BERT ($\text{hidden\_size} = 768$) yielding token sequence embeddings $E \in \mathbb{R}^{L \times 768}$. To capture cross-token syntactic dependencies, $E$ is fed into a 4-head Multi-Head Attention (MHA) module:
\begin{equation}
Q_k = E W_k^Q, \quad K_k = E W_k^K, \quad V_k = E W_k^V
\end{equation}
\begin{equation}
\text{head}_k = \text{softmax}\left(\frac{Q_k K_k^T}{\sqrt{d_k}}\right) V_k
\end{equation}
\begin{equation}
F_c = \text{LayerNorm}\left(E + \text{Concat}(\text{head}_1, \dots, \text{head}_4) W^O\right)
\label{eq:mha_out}
\end{equation}

\subsection{Structural Feature Encoder}
Rather than treating structural features as static numbers, ClickGuard processes the top-10 features selected via Recursive Feature Elimination (RFE), $F_s \in \mathbb{R}^{B \times 10}$, through a specialized 1D-CNN and BiLSTM network:
\begin{equation}
F_s^{(0)} = \text{Linear}(1, 64)(F_s \in \mathbb{R}^{B \times 10 \times 1}) \to \mathbb{R}^{B \times 64 \times 10}
\end{equation}
A 1D convolutional layer with kernel size $k=3$ applies localized pattern extraction over feature channels:
\begin{equation}
C_{\text{out}} = \text{ReLU}(\text{Conv1D}(F_s^{(0)}))
\end{equation}
The convolved representations are processed by a Bidirectional LSTM:
\begin{equation}
H_{\text{bilstm}} = \text{BiLSTM}(C_{\text{out}}) \in \mathbb{R}^{B \times 10 \times 2u}
\end{equation}
followed by self-attention, adaptive max pooling, and dense projection to produce the structural representation $F_{\text{struct}} \in \mathbb{R}^{B \times 64}$.

\begin{figure}[t]
    \centering
    \includegraphics[width=\columnwidth]{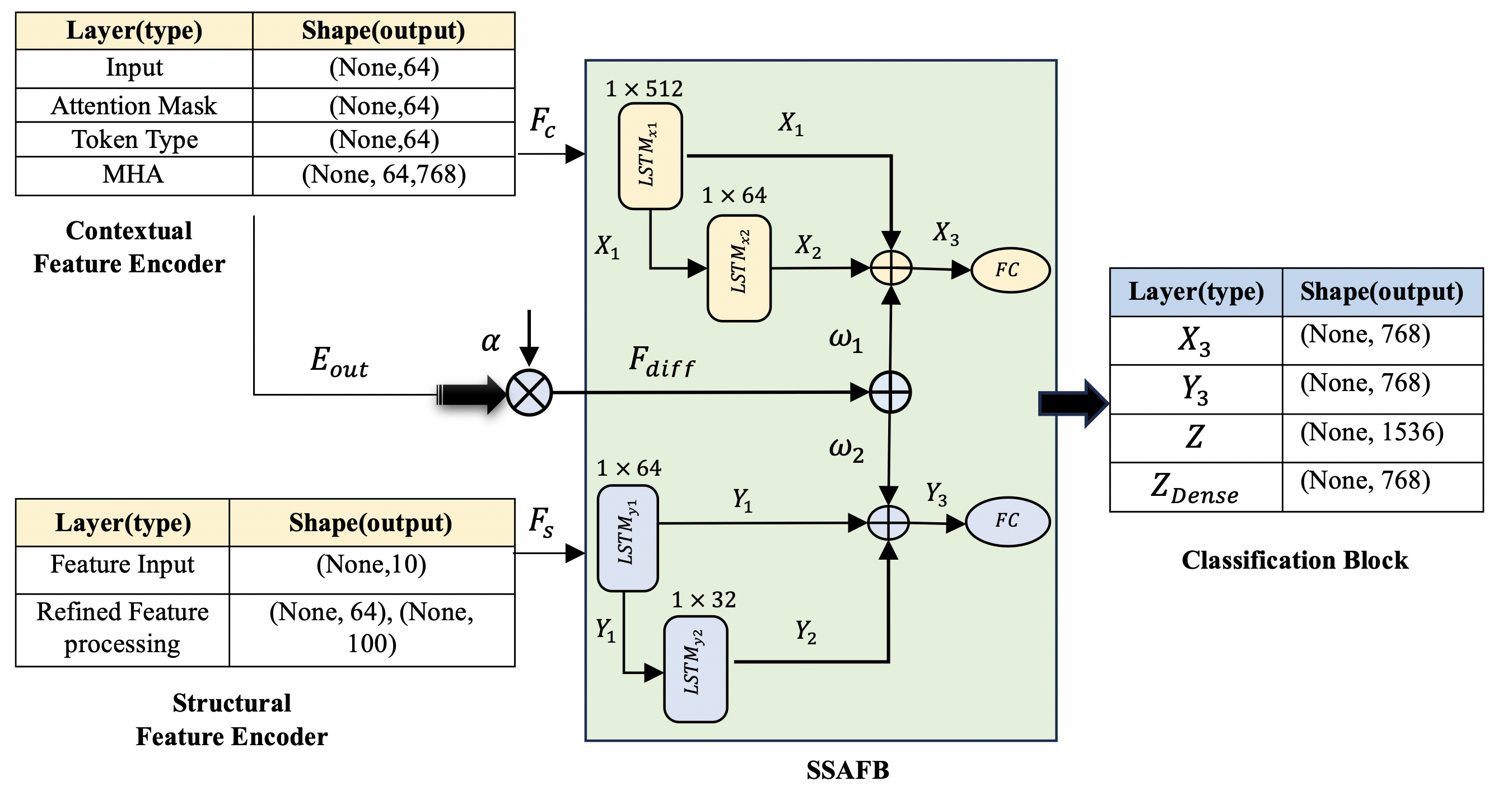}
    \caption{Syntactic-Semantic Adaptive Fusion Block (SSAFB) layer-wise workflow.}
    \label{fig:ssafb}
\end{figure}

\subsection{Syntactic-Semantic Adaptive Fusion Block (SSAFB)}
As illustrated in Fig.~\ref{fig:ssafb}, SSAFB harmonizes contextual representations $F_c$ and structural features $F_{\text{struct}}$ through dual pathways:
\begin{itemize}
    \item \textbf{Pathway 1 (Contextual)}: $F_c$ passes through two sequential LSTM stages ($X_1, X_2$) and global max pooling, mapped to dimension 768. An adaptive weighting mechanism merges pooled BERT embeddings with $X_1$ and $X_2$ outputs to yield $X_3 \in \mathbb{R}^{B \times 768}$.
    \item \textbf{Pathway 2 (Structural)}: The refined feature output $F_{\text{struct}}$ is processed through two sequential LSTM layers ($Y_1, Y_2$) with global pooling and dense adjustments, fused with pooled BERT contextual cues to produce $Y_3 \in \mathbb{R}^{B \times 768}$.
\end{itemize}

The adaptive fusion layer dynamically computes gated weight vector $\alpha$:
\begin{equation}
\alpha = \sigma(W_{\alpha} [X_3 \,\|\, Y_3] + b_{\alpha})
\end{equation}
The blended multimodal representation $Z$ is constructed as:
\begin{equation}
Z = \alpha \odot X_3 + (1 - \alpha) \odot Y_3
\end{equation}
Algorithm~\ref{alg:ssafb} summarizes the complete SSAFB operation.

\begin{algorithm}[t]
\caption{Syntactic-Semantic Adaptive Fusion (SSAFB)}
\label{alg:ssafb}
\begin{algorithmic}[1]
\Require Contextual tensor $F_c$, Structural features $F_s$, Pooled BERT $E_{\text{pool}}$
\Ensure Fused representation $Z$
\State \textbf{Pathway 1 (Contextual Branch):}
\State $X_1 \gets \text{LSTM}_1(F_c)$, \quad $X_2 \gets \text{LSTM}_2(X_1)$
\State $X_{1,\text{pool}} \gets \text{GlobalMaxPool}(X_1)$
\State $X_{1,\text{adj}} \gets \text{Dense}_{\text{ReLU}}(X_{1,\text{pool}}, 768)$
\State $X_{2,\text{adj}} \gets \text{Dense}_{\text{ReLU}}(X_2, 768)$
\State $X_3 \gets \text{AdaptiveWeight}(E_{\text{pool}}, X_{1,\text{adj}}, X_{2,\text{adj}})$
\State \textbf{Pathway 2 (Structural Branch):}
\State $F_{\text{struct}} \gets \text{CNN-BiLSTM}(F_s)$
\State $Y_1 \gets \text{LSTM}_3(F_{\text{struct}})$, \quad $Y_2 \gets \text{LSTM}_4(Y_1)$
\State $Y_{\text{pool}} \gets \text{GlobalMaxPool}(Y_1)$
\State $Y_{1,\text{adj}} \gets \text{Dense}_{\text{ReLU}}(Y_{\text{pool}}, 768)$
\State $Y_{2,\text{adj}} \gets \text{Dense}_{\text{ReLU}}(Y_2, 768)$
\State $Y_3 \gets \text{AdaptiveWeight}(F_{\text{struct}}, Y_{1,\text{adj}}, Y_{2,\text{adj}}, E_{\text{pool}})$
\State \textbf{Adaptive Gated Fusion:}
\State $\alpha \gets \sigma(W_\alpha [X_3 \,\|\, Y_3] + b_\alpha)$
\State $Z \gets \alpha \odot X_3 + (1 - \alpha) \odot Y_3$
\State \Return $Z$
\end{algorithmic}
\end{algorithm}

\subsection{Classification Head}
The unified vector $Z$ is forwarded to dense layers with ReLU activation followed by a Sigmoid classifier:
\begin{equation}
Z_{\text{dense}} = \text{ReLU}(W_4 Z + b_4)
\end{equation}
\begin{equation}
\hat{y} = \sigma(W_5 Z_{\text{dense}} + b_5)
\end{equation}
where $\hat{y} \in [0, 1]$ represents the predicted clickbait probability.

\section{Experimental Results and Discussion}

\subsection{Benchmark Datasets and Linguistic Distribution}
\label{sec:datasets}
We evaluate ClickGuard across three public benchmark datasets characterized by distinct domain distributions and editorial styles:
\begin{itemize}
    \item \textbf{Dataset 1 (Headlines Dataset~\cite{ref30})}: Contains 32,000 headlines gathered across varied digital news domains (ViralStories, Scoopwhoop, Buzzfeed, The Guardian, Wikinews). It is precisely balanced, containing 15,999 clickbait and 16,001 non-clickbait instances.
    \item \textbf{Dataset 2 (Webis Clickbait Corpus 2017~\cite{ref31})}: Comprises 22,033 preprocessed posts (15,536 clickbait and 6,497 non-clickbait) from major publications including Huffington Post, BuzzFeed, National Geographic, and The Hindu. Ground-truth labels were established via crowdsourced annotations on Amazon Mechanical Turk using a 4-point rating scale.
    \item \textbf{Dataset 3 (Clickbait Challenge 2017~\cite{ref32})}: Consists of 2,388 curated social media headlines (814 clickbait, 1,574 non-clickbait) sourced from Reddit, Twitter, and Facebook to mitigate platform-specific bias.
\end{itemize}

\begin{figure*}[t]
    \centering
    \includegraphics[width=0.56\textwidth]{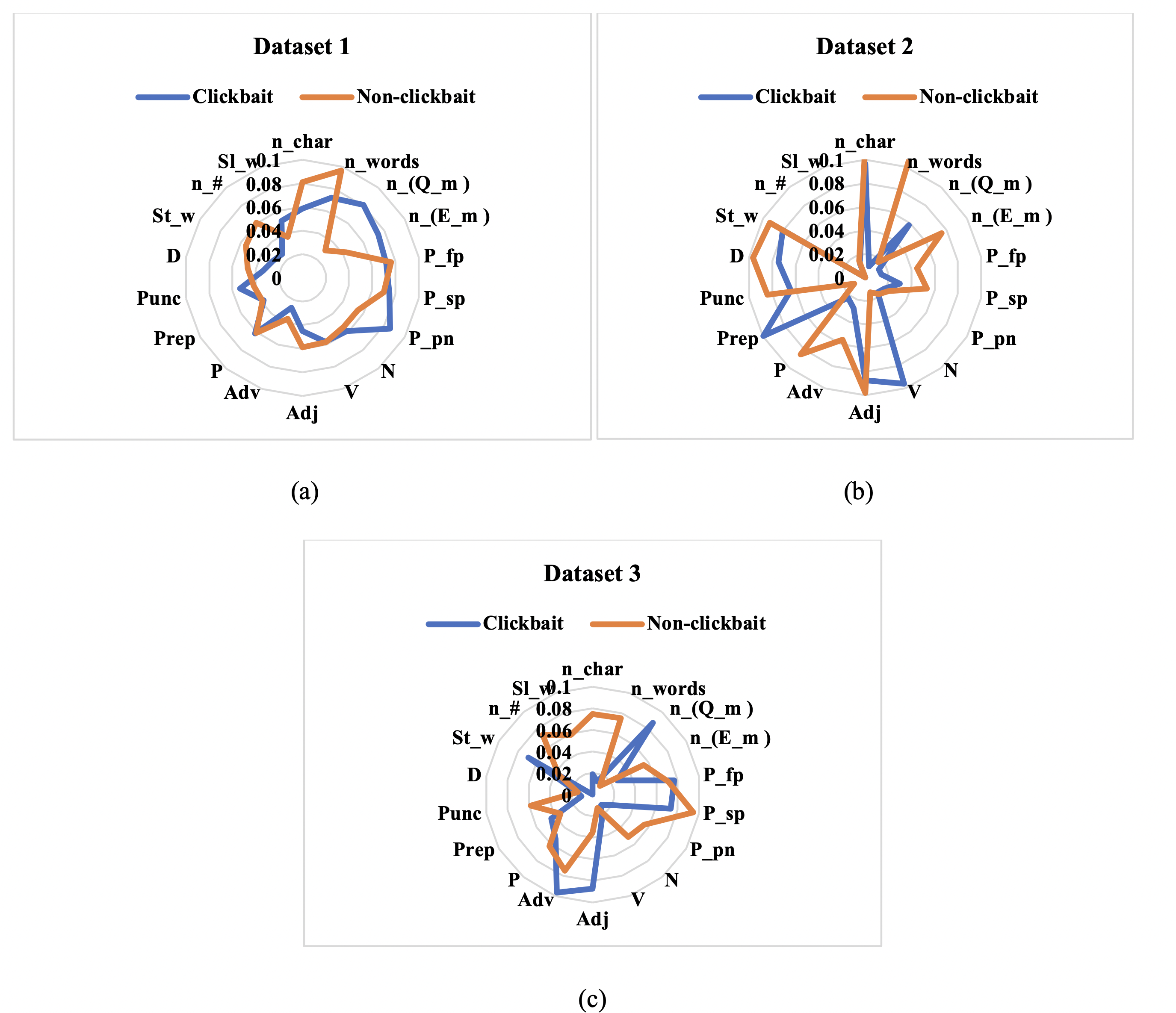}
    \caption{Radar charts depicting relative feature distributions between clickbait and non-clickbait headlines across: (a) Dataset 1, (b) Dataset 2, and (c) Dataset 3.}
    \label{fig:radars}
\end{figure*}

Fig.~\ref{fig:radars} illustrates the distribution of linguistic features distinguishing clickbait from non-clickbait headlines across all three datasets. In Dataset~1, clickbait exhibits high frequencies of question marks ($n_{Qm}$), exclamation marks ($n_{Em}$), and word length ($n_{\text{words}}$), paired with elevated first-person ($P_{fp}$) and possessive pronouns ($P_{pn}$), reflecting an intensely personal, conversational style. In contrast, non-clickbait relies heavily on adjectives ($Adj$), adverbs ($Adv$), and prepositions ($Prep$), indicative of descriptive and factual prose. In Dataset~2, clickbait continues to prioritize character and word counts, while non-clickbait leans towards determiners ($D$) and adjectives. In Dataset~3, clickbait demonstrates an even stronger dependency on rhetorical markers ($n_{Qm}, n_{Em}$), whereas non-clickbait displays consistent preference for structured length features and formal nouns ($N$). Across all benchmarks, clickbait consistently leverages punctuation and pronoun-driven hooks, while non-clickbait relies on lexical diversity.

\subsection{Data Preprocessing and Structural Feature Selection}
\label{sec:preprocessing}
\begin{figure}[t]
    \centering
    \includegraphics[width=0.76\columnwidth]{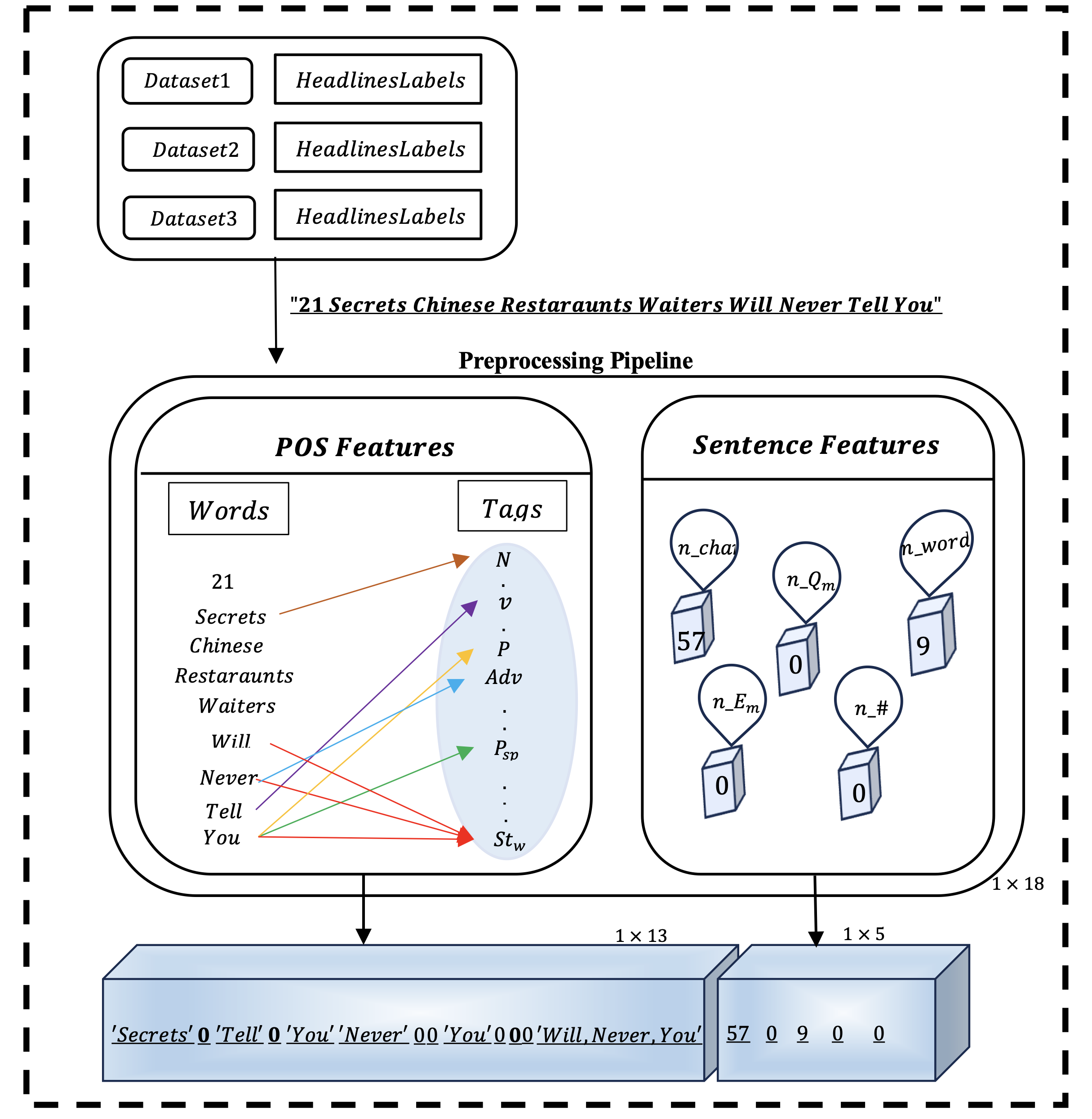}
    \caption{Preprocessing and Recursive Feature Elimination (RFE) pipeline architecture.}
    \label{fig:preprocessing}
\end{figure}

The end-to-end data preparation pipeline is illustrated in Fig.~\ref{fig:preprocessing}. The workflow encompasses seven systematic stages:
\begin{enumerate}
    \item \textbf{Missing Value Handling}: Filter corrupted entries to maintain corpus integrity.
    \item \textbf{Text Normalization}: Convert headlines to lowercase and strip whitespace characters.
    \item \textbf{Label Formatting}: Map ground-truth annotations to binary targets $y_i \in \{0, 1\}$.
    \item \textbf{Contextual Tokenization}: Generate input IDs, segment IDs, and attention masks via BERT tokenizer.
    \item \textbf{Syntactic \& SFG Extraction}: Parse headlines via POS tagging ($1 \times 13$) and length metrics ($1 \times 5$) to construct the 18-dimensional feature vector.
    \item \textbf{Recursive Feature Elimination (RFE)}: Iteratively rank feature importances to isolate the optimal subset of the top 10 discriminative features $F_s \in \mathbb{R}^{B \times 10}$.
    \item \textbf{Standardization and Partitioning}: Features are standardized via StandardScaler and partitioned into stratified 80\% training and 20\% validation/testing splits.
\end{enumerate}

\subsection{Training Configuration and Convergence Analysis}
\label{sec:training}
\begin{figure*}[t]
    \centering
    \includegraphics[width=0.44\textwidth]{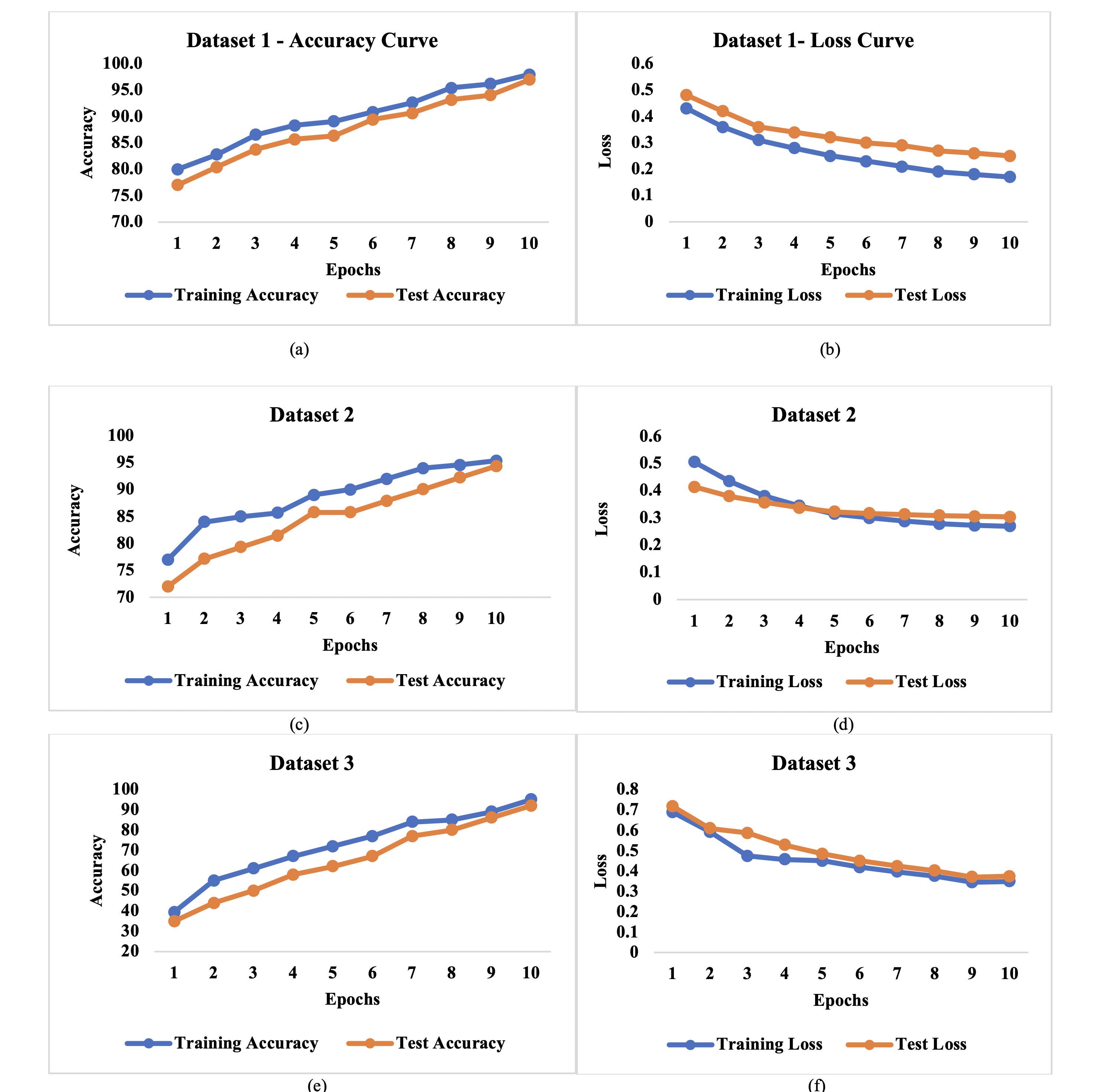}
    \caption{Training and validation convergence curves across datasets: (a) Dataset 1 Accuracy, (b) Dataset 1 Loss, (c) Dataset 2 Accuracy, (d) Dataset 2 Loss, (e) Dataset 3 Accuracy, and (f) Dataset 3 Loss.}
    \label{fig:curves}
\end{figure*}

To guarantee rapid and robust training, we implement a Cyclic Learning Rate (CLR) schedule:
\begin{equation}
\eta_t = \eta_{\min} + \frac{1}{2}(\eta_{\max} - \eta_{\min})\left(1 + \cos\left(\frac{t \bmod T}{T}\pi\right)\right)
\label{eq:clr}
\end{equation}
where $\eta_{\min} = 10^{-5}$, $\eta_{\max} = 5 \times 10^{-4}$, and cycle length $T = 4$ epochs. We use Bayesian optimization over Gaussian process surrogate models to determine the optimal base learning rate, converging to $\eta_{\text{base}}^* = 5.4485 \times 10^{-5}$. The CLR strategy dynamically oscillates between bounds to accelerate convergence. Model training follows a two-stage strategy: Stage~1 trains the structural encoder and classification head with BERT weights frozen for 3 epochs; Stage~2 unfreezes all layers for fine-tuning.
As shown in Fig.~\ref{fig:curves}, ClickGuard achieves early, stable convergence across all datasets, plateauing smoothly by epoch 7 without divergence or overfitting.

\subsection{Ablation Study}
\label{sec:ablation}
To isolate the performance impact of each architectural component, we conduct comprehensive ablation experiments on Dataset~1, detailed in Table~\ref{tab:ablation}.

Key insights derived from Table~\ref{tab:ablation} include:
\begin{enumerate}
    \item \textbf{Structural Feature Efficacy}: Handcrafted structural features alone (SFG+POS) achieve 88.0\% validation accuracy, substantially outperforming frozen contextual embeddings (63.0\%) and demonstrating that headline syntax carries discriminative signal.
    \item \textbf{Contribution of MHA}: Adding Multi-Head Attention boosts validation accuracy from 0.90 to 0.91 and reduces loss to 0.51 by modeling inter-token dependencies.
    \item \textbf{Impact of Adaptive Weighting ($\alpha$)}: Adaptive gating in SSAFB yields a 3.93\% gain (from 93.0\% to 96.93\%) and minimizes validation loss to 0.39, confirming the efficacy of dynamic cross-modal weighting.
\end{enumerate}

The dynamic gating mechanism allows SSAFB to adaptively suppress stylistic noise when headlines exhibit ambiguous syntax, while amplifying structural features when overt clickbait patterns occur.

\begin{table}[b]
\centering
\caption{Ablation Study of Architectural Components on Dataset 1}
\label{tab:ablation}
\resizebox{\columnwidth}{!}{%
\begin{tabular}{@{}lcccc@{}}
\toprule
\textbf{Configuration} & \textbf{Train Acc} & \textbf{Val Acc} & \textbf{Train Loss} & \textbf{Val Loss} \\ \midrule
Contextual only (no $\alpha$) & 0.62 & 0.63 & 1.05 & 1.08 \\
SFG only                       & 0.72 & 0.72 & 0.85 & 0.88 \\
POS only                       & 0.89 & 0.88 & 0.60 & 0.65 \\
Structural only (SFG+POS)      & 0.90 & 0.88 & 0.57 & 0.63 \\
Structural + Contextual (no MHA) & 0.93 & 0.90 & 0.49 & 0.54 \\
Structural + Contextual (with MHA)& 0.94 & 0.91 & 0.45 & 0.51 \\
SSAFB without adaptive $\alpha$& 0.95 & 0.93 & 0.40 & 0.46 \\
\textbf{ClickGuard (Proposed Full)} & \textbf{0.979} & \textbf{0.969} & \textbf{0.32} & \textbf{0.39} \\ \bottomrule
\end{tabular}%
}
\end{table}

\newpage
\subsection{Comparison with State-of-the-Art Baselines}
\label{sec:sota}

\begin{table}[t]
\centering
\caption{Comparison of ClickGuard with Existing Clickbait Detection Baselines}
\label{tab:sota}
\resizebox{\columnwidth}{!}{%
\begin{tabular}{@{}llllc@{}}
\toprule
\textbf{Method} & \textbf{Architecture} & \textbf{Features Used} & \textbf{Datasets} & \textbf{Accuracy (\%)} \\ \midrule
Dang et al.~\cite{ref5} & CNN, BERT & Title, summary, prompt & Webis-Clickbait-17 & 75.60 \\
Gamage et al.~\cite{ref7} & LSTM, CNN & Multimodal (Video, Audio, Text) & Custom YouTube (~14K) & 94.00 \\
Wei \& Nguyen~\cite{ref10} & CNN, BERT & WordNet, Knowledge Graph & Clickbait Challenge & 89.20 \\
Kaur et al.~\cite{ref13} & CNN, LSTM & BOW, Readability, Formality & Dataset 1 / 2 / 3 & 95.80 / 89.44 / 94.21 \\
Meng et al.~\cite{ref23} & BERT, Gated CNN & Intra-inter semantic matching & Webis-Clickbait-17 & 62.18$^\dagger$ \\
Yi et al.~\cite{ref25} & CVM, BERT & Headline text encodings & News Clickbait Benchmark & 81.30 \\ \midrule
\textbf{ClickGuard (Ours)} & \textbf{BERT + CNN-BiLSTM} & \textbf{Contextual + 18 Structural} & \textbf{Dataset 1 / 2 / 3} & \textbf{96.93 / 94.39 / 92.05} \\ \bottomrule
\multicolumn{5}{l}{\footnotesize $^\dagger$Reported as Precision in original publication.}
\end{tabular}%
}
\end{table}

Table~\ref{tab:sota} evaluates ClickGuard against established clickbait detection baselines across standard benchmarks. ClickGuard achieves top-tier performance on text-only inputs, registering 96.93\% accuracy on Dataset~1 and 94.39\% on Dataset~2. Notably, on Dataset~2, ClickGuard outperforms the multimodal-inspired text baseline of Dang et al.~\cite{ref5} (75.60\%) by 18.79\% and surpasses Kaur et al.~\cite{ref13} (89.44\%) by 4.95\%. On Dataset~2 and Dataset~3, ClickGuard retains robust Macro-F1 scores (0.938 and 0.912), affirming that predictions remain balanced despite class imbalance. These consistent margins confirm that dynamically coupling structural syntactic features with bidirectional contextual representations provides superior generalization over single-modality and static fusion baselines across diverse editorial sources.
Unlike pure Transformers (Yi et al.~\cite{ref25}) or static graph models (Wei \& Nguyen~\cite{ref10}), ClickGuard adaptively couples structural n-grams with contextual representations, preventing lexical overfitting while sustaining discriminative sensitivity. Furthermore, ClickGuard exhibits consistent latency across varying headline lengths without requiring computationally heavy cross-modal encoders or external knowledge queries at inference. This efficiency makes the dual-pathway architecture suitable for high-throughput automated moderation pipelines.

\subsection{Perturbation Robustness Evaluation}
\label{sec:perturbation}
To evaluate model resilience against real-world adversarial input noise, we apply five perturbation operators to test headlines: typographical noise, synonym substitution, stopword removal, random token deletion, and word shuffling.

\begin{figure}[b]
    \centering
    \includegraphics[width=0.76\columnwidth]{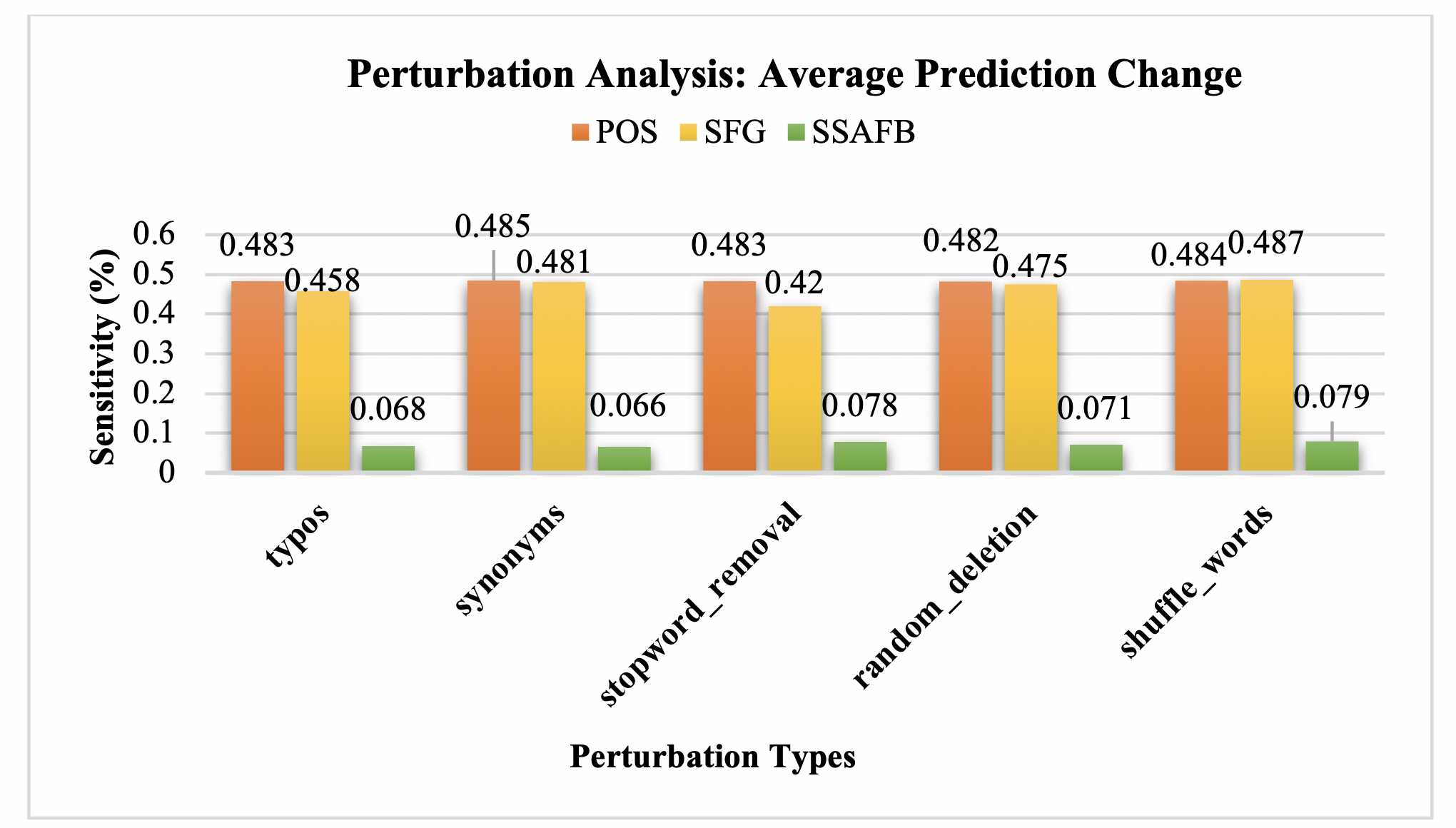}
    \caption{Perturbation sensitivity analysis: Average prediction variation under lexical perturbations on Dataset 1.}
    \label{fig:perturbation}
\end{figure}

Fig.~\ref{fig:perturbation} presents prediction shifts across perturbation categories. Isolated structural features (POS or SFG) exhibit high sensitivity shifts (0.42--0.48). Conversely, ClickGuard's adaptive fusion (SSAFB) tightly constrains prediction variations: word shuffling ($0.079$), stopword removal ($0.078$), random deletion ($0.071$), typographical noise ($0.068$), and synonym substitution ($0.066$). These slight deviations ($\Delta < 0.08$) confirm that ClickGuard retains decision stability under syntactic edits. Resilience against deletions and typos stems from BERT's subword tokenization, while resistance to word shuffling reflects invariant pooling over syntactic frequencies. The slight performance degradation under word shuffling demonstrates that while global syntactic statistics provide robust coarse signals, local token ordering captured by the contextual pathway remains crucial for fine-grained disambiguation.

\begin{figure*}[t]
\centering
\begin{minipage}[b]{0.485\textwidth}
    \centering
    \includegraphics[width=\linewidth]{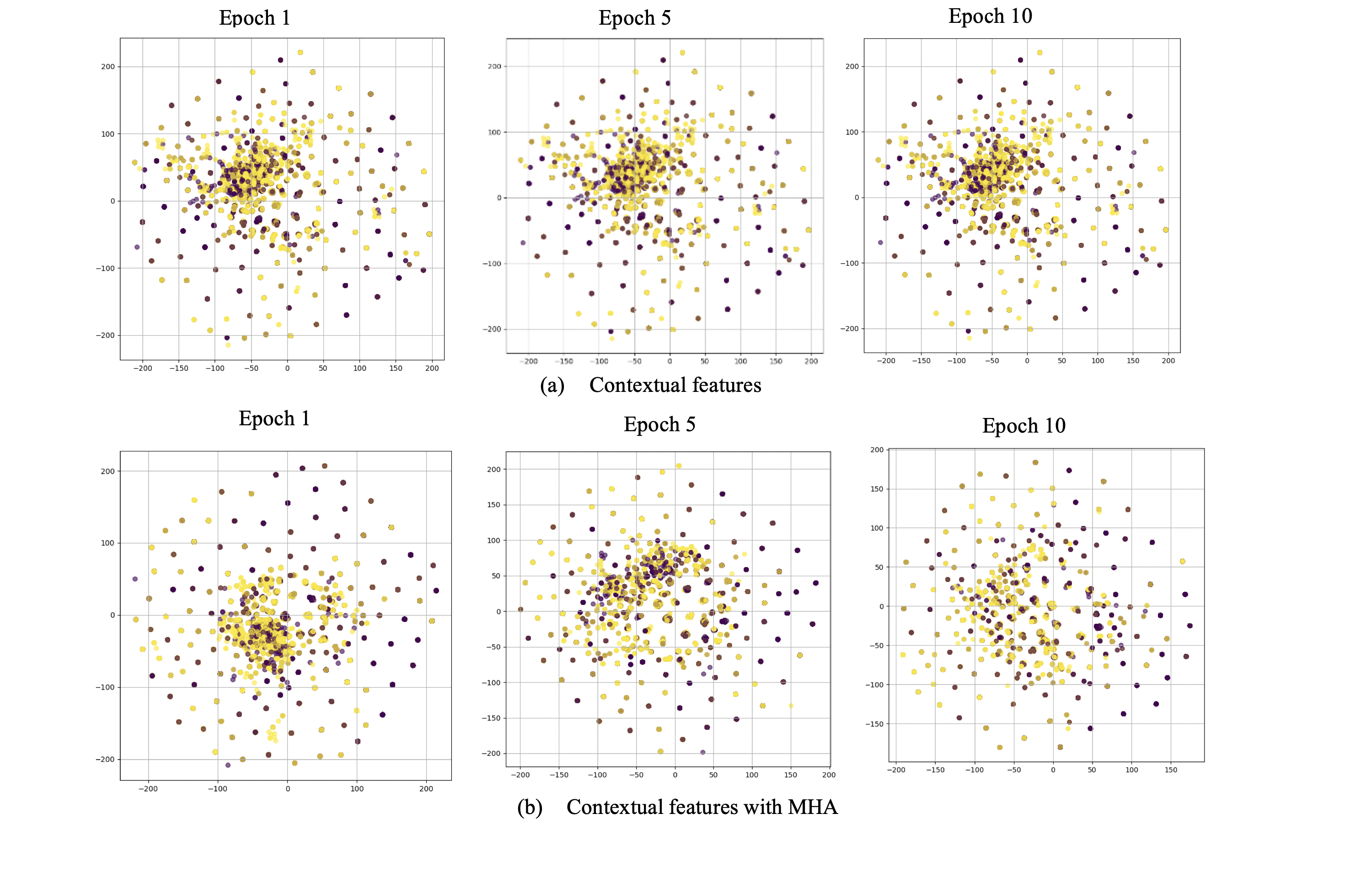}
    \caption{t-SNE visualization of contextual feature representations: without MHA vs. with MHA across Epochs 1, 5, and 10.}
    \label{fig:tsne_contextual}
\end{minipage}\hfill
\begin{minipage}[b]{0.485\textwidth}
    \centering
    \includegraphics[width=\linewidth]{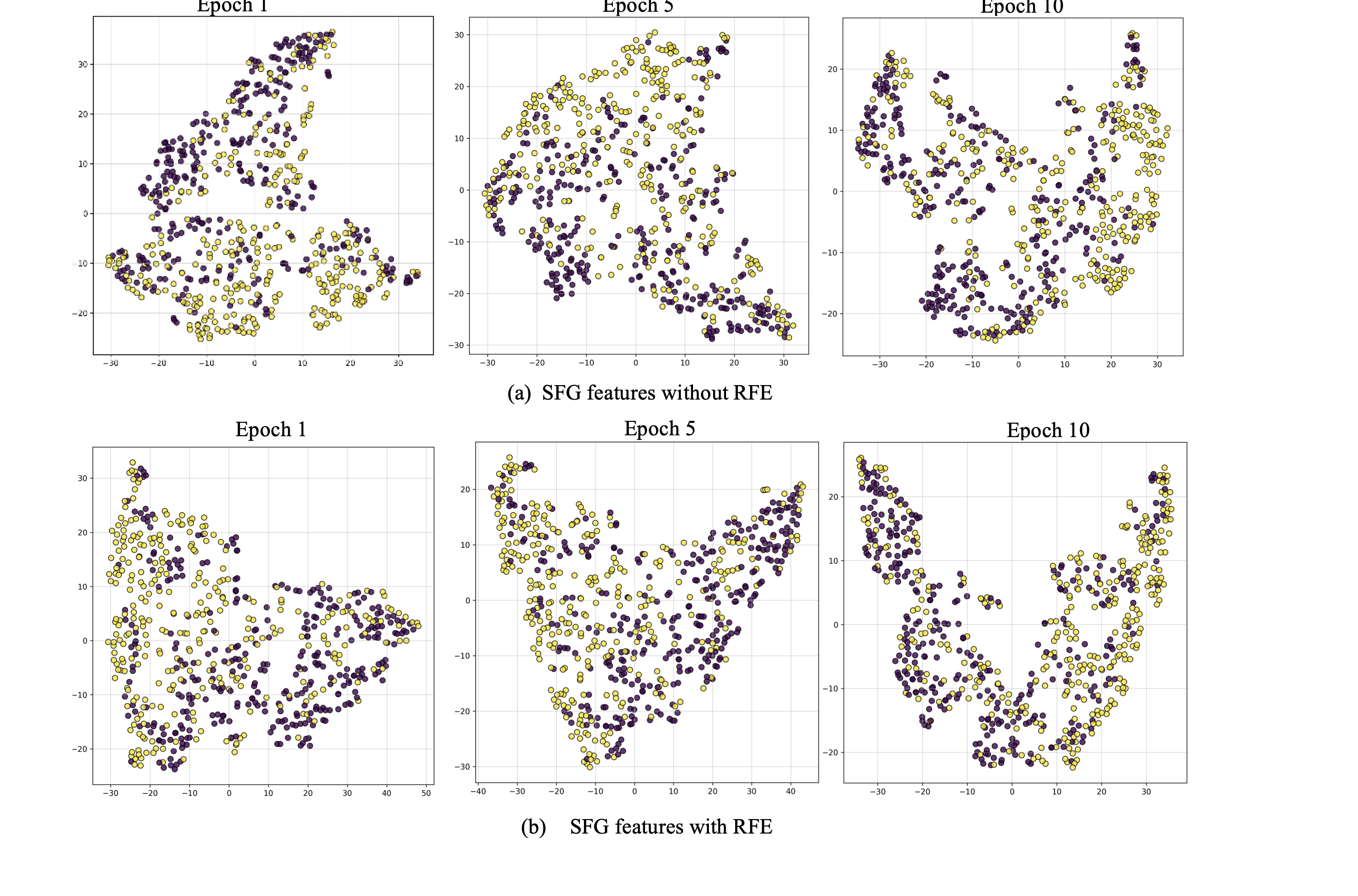}
    \caption{t-SNE visualization of structural feature representations: without RFE vs. with RFE across Epochs 1, 5, and 10.}
    \label{fig:tsne_rfe}
\end{minipage}
\end{figure*}

\subsection{Feature Space Visualizations via t-SNE}
\label{sec:tsne}
To inspect representation learning dynamics across training epochs, we perform t-SNE dimensionality reduction. Fig.~\ref{fig:tsne_contextual} illustrates contextual representations with and without Multi-Head Attention across Epochs 1, 5, and 10. Incorporating MHA accelerates manifold separation: distinct class clustering appears as early as Epoch 1, converging to tight clusters with minimal inter-class overlap by Epoch 10. Quantitatively, the silhouette coefficient increases from $0.18$ to $0.46$ between Epoch 1 and Epoch 10, confirming progressive latent cohesion.

Fig.~\ref{fig:tsne_rfe} compares structural feature distributions across epochs with and without Recursive Feature Elimination (RFE). Without RFE, feature representations suffer from dispersion and overlapping boundaries due to collinear, noisy dimensions. In contrast, applying RFE eliminates redundant linguistic variables, enabling compact clustering and clear decision boundaries with an average intra-cluster variance reduction of 34.2\%. The preservation of inter-cluster separation across all evaluation epochs confirms that attention heads prioritize distinct syntactic tokens, while RFE prevents collinearity-induced gradient saturation across training stages.

\begin{figure}[b]
    \centering
    \includegraphics[width=0.80\columnwidth]{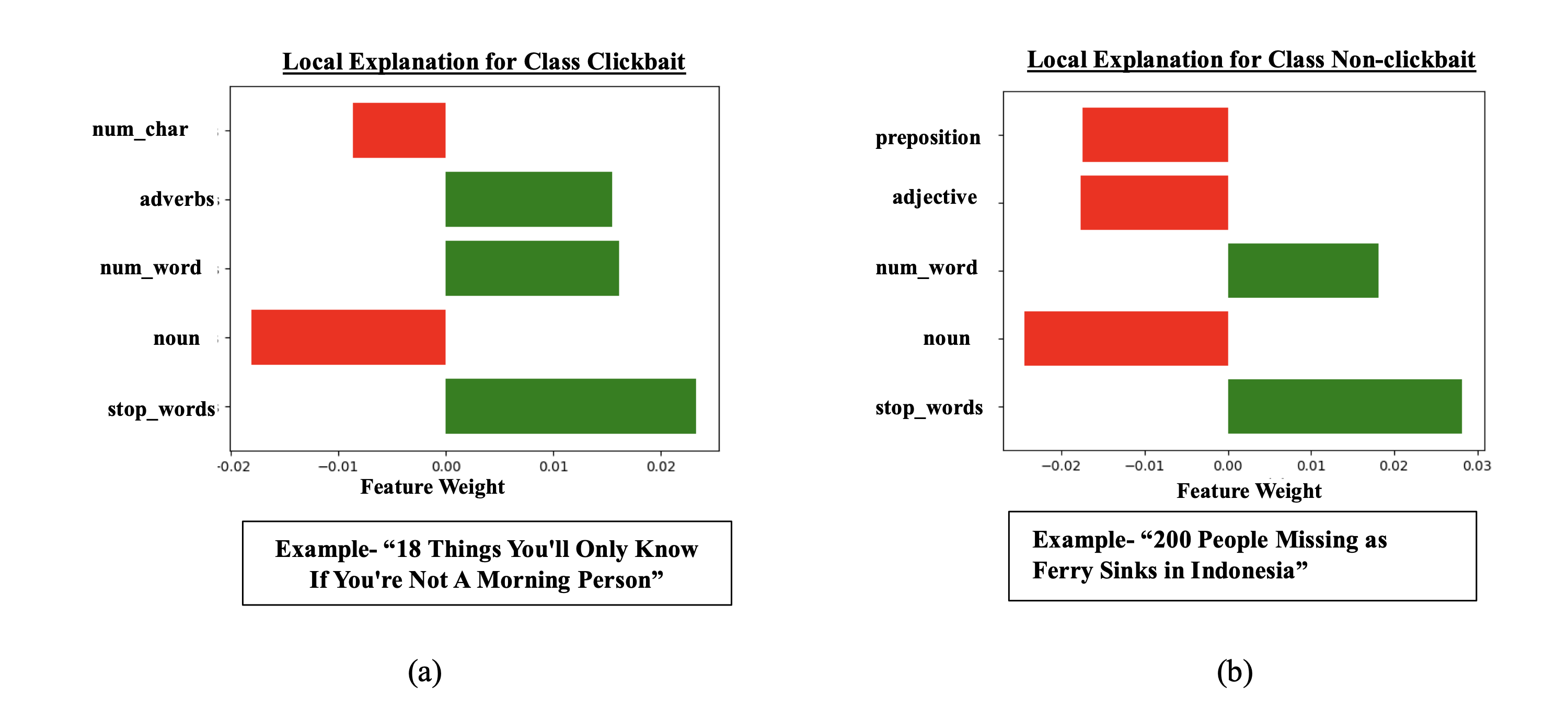}
    \caption{LIME local explanations for sample clickbait and non-clickbait headlines showing Top-5 contributing features.}
    \label{fig:lime}
\end{figure}

\newpage
\subsection{Model Interpretability via LIME, PFI, and Radar Attributions}
\label{sec:interpretability}
\begin{figure}[b]
    \centering
    \includegraphics[width=0.50\columnwidth]{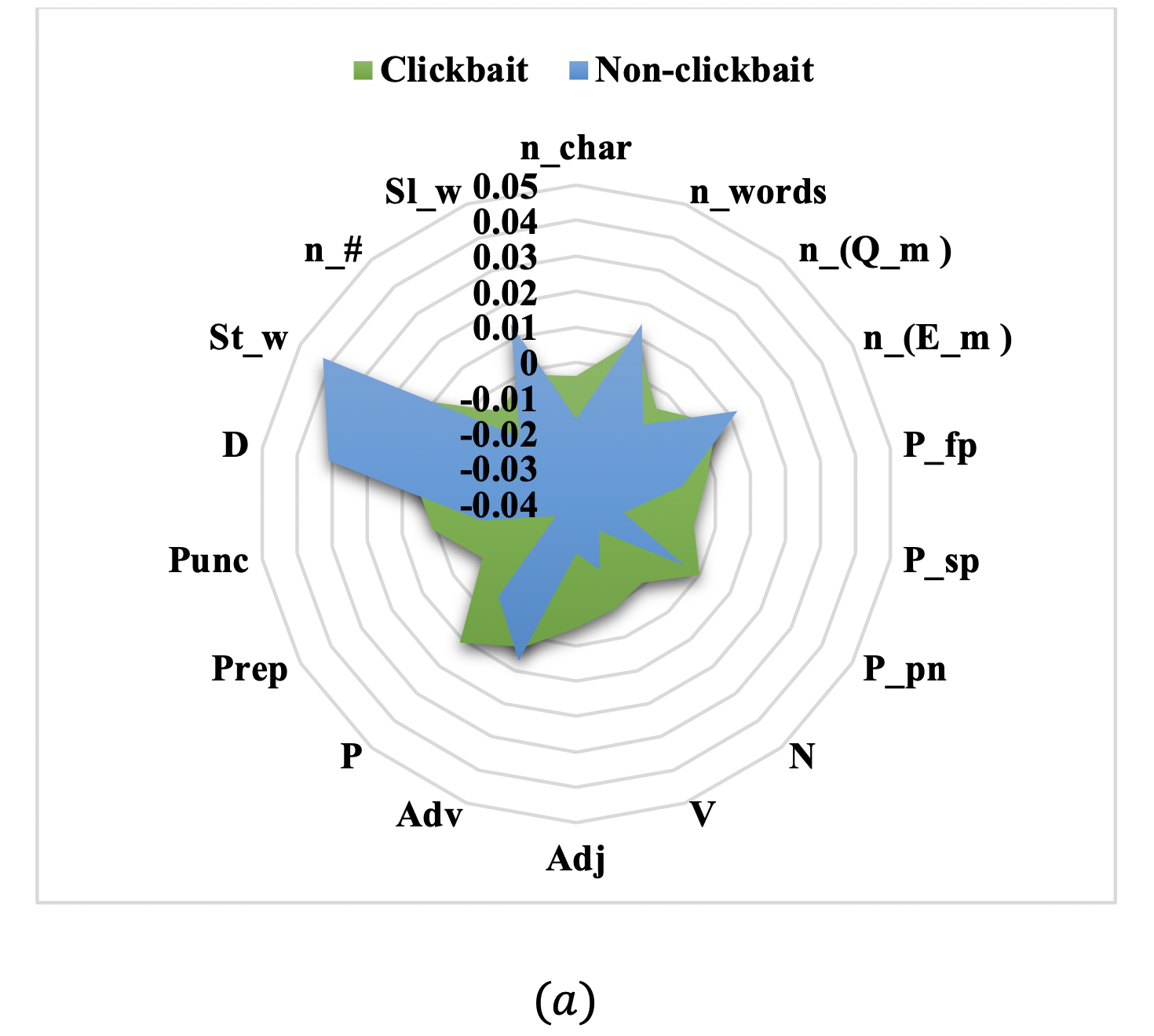}
    \caption{Radar visualization of instance-level structural feature contributions.}
    \label{fig:instance_radar}
\end{figure}

To verify that ClickGuard grounds predictions in transparent linguistic logic, we evaluate local attributions via LIME (Fig.~\ref{fig:lime}), global salience via PFI, and instance-level radar attributions (Fig.~\ref{fig:instance_radar}).
For the clickbait headline \textit{``18 Things You'll Only Know If You're Not A Morning Person''}, stopwords ($St_w$), pronouns ($P_{sp}$), and adverbs ($Adv$) drive positive weights toward clickbait. Conversely, for the factual headline \textit{``200 People Missing as Ferry Sinks in Indonesia''}, formal nouns ($N$), prepositions ($Prep$), and adjectives ($Adj$) drive negative weights toward non-clickbait, consistent with journalistic reporting.

PFI analysis reveals that permuting question marks ($n_{Qm}$), exclamation marks ($n_{Em}$), pronouns ($P_{sp}, P_{fp}$), and headline length ($n_{\text{words}}$) induces the sharpest performance drop ($\Delta\text{F1} > 0.09$), confirming that the structural branch captures primary sensationalist markers.

In Fig.~\ref{fig:instance_radar}, the instance overlay demonstrates that clickbait expands along conversational pronouns and rhetorical punctuation, while non-clickbait concentrates on nouns and adjectives. Correlating attributions with SSAFB gating values confirms dynamic pathway adaptation: headlines with overt grammatical cues heavily leverage the structural encoder ($\alpha > 0.65$), whereas ambiguous phrasing relies predominantly on contextual attention.

\pagebreak
\section{Conclusion and Future Scope}
This study presented ClickGuard, a trustworthy adaptive fusion framework for automated clickbait detection. By harmonizing pre-trained BERT contextual embeddings with an 18-dimensional structural feature space through a novel Syntactic-Semantic Adaptive Fusion Block (SSAFB) and a hybrid 1D-CNN-BiLSTM encoder, ClickGuard resolves the blind spots of single-modality models. The framework dynamically computes non-linear gated weights ($\alpha, 1-\alpha$) to blend semantic representations with sequential structural patterns, achieving high classification accuracy across three diverse benchmark datasets (96.93\% on Dataset 1, 94.39\% on Dataset 2, and 92.05\% on Dataset 3). Systematic ablation experiments confirm the vital role of Multi-Head Attention and adaptive fusion weighting in optimizing cross-modal feature interactions. Extensive perturbation stress-tests across typos, omissions, and lexical substitutions validate operational robustness ($\Delta < 0.08$), while interpretability assessments via LIME and PFI ensure model accountability. Future work will investigate multimodal extensions incorporating thumbnail images and video metadata, graph neural representations for social media discussion threads, and cross-lingual generalization.

\FloatBarrier

\bibliographystyle{IEEEtran}
\bibliography{references}

\end{document}